\documentclass[conference]{IEEEtran}
\makeatletter
\def\ps@headings{%
\def\@oddhead{\mbox{}\scriptsize\rightmark \hfil \thepage}%
\def\@evenhead{\scriptsize\thepage \hfil \leftmark\mbox{}}%
\def\@oddfoot{}%
\def\@evenfoot{}}
\makeatother 

\usepackage{cite}      % Written by Donald Arseneau
\usepackage{subfigure} % Written by Steven Douglas Cochran
\usepackage{url}       % Written by Donald Arseneau
\usepackage{algorithm}
\usepackage{array}
\usepackage{amsmath,amsfonts,amssymb,graphicx}
\usepackage{bm}
\usepackage{mathrsfs}
\usepackage{multicol}
\usepackage[end]{algpseudocode}
\usepackage{multirow}
\usepackage{dsfont}

\newtheorem{theorem}{\textbf{Theorem}}
\newtheorem{lemma}{\textbf{Lemma}}

\newtheorem{remark}{\textbf{Remark}}

\newcommand {\xA} {\mathcal{A}}

\newcommand{\commnt}[1] {$//$ \textsc{#1} }
\newcommand{\xT} {\widetilde{T}_{i}}
\newcommand{\xI} {\widetilde{I}_{i}}
\newcommand{\xF} {\mathcal{F}}
\newcommand {\xa} {\mathbf{a}}

\newcommand{\aX} {\overline{X}}

\begin{document}
\title{Online Learning Algorithms for Stochastic Water-Filling}

\author{\IEEEauthorblockN{Yi Gai and Bhaskar Krishnamachari}
\IEEEauthorblockA{Ming Hsieh Department of Electrical Engineering\\
University of Southern California\\
Los Angeles, CA 90089, USA\\
Email: $\{$ygai, bkrishna$\}$@usc.edu}}

\maketitle

\begin{abstract}
Water-filling is the term for the classic solution to the problem of
allocating constrained power to a set of parallel channels to
maximize the total data-rate. It is used widely in practice, for
example, for power allocation to sub-carriers in multi-user OFDM
systems such as WiMax. The classic water-filling algorithm is
deterministic and requires perfect knowledge of the channel gain to
noise ratios. In this paper we consider how to do power allocation
over stochastically time-varying (i.i.d.) channels with unknown gain
to noise ratio distributions. We adopt an online learning framework
based on stochastic multi-armed bandits. We consider two variations
of the problem, one in which the goal is to find a power allocation
to maximize $\sum\limits_i \mathbb{E}[\log(1 + SNR_i)]$, and another
in which the goal is to find a power allocation to maximize
$\sum\limits_i \log(1 + \mathbb{E}[SNR_i])$. For the first problem,
we propose a \emph{cognitive water-filling} algorithm that we call
CWF1. We show that CWF1 obtains a regret (defined as the cumulative
gap over time between the sum-rate obtained by a distribution-aware
genie and this policy) that grows polynomially in the number of
channels and logarithmically in time, implying that it
asymptotically achieves the optimal time-averaged rate that can be
obtained when the gain distributions are known. For the second
problem, we present an algorithm called CWF2, which is, to our
knowledge, the first algorithm in the literature on stochastic
multi-armed bandits to exploit non-linear dependencies between the
arms. We prove that the number of times CWF2 picks the incorrect
power allocation is bounded by a function that is polynomial in the
number of channels and logarithmic in time, implying that its
frequency of incorrect allocation tends to zero.
\end{abstract}

\section{Introduction}\label{sec:intro}

A fundamental resource allocation problem that arises in many
settings in communication networks is to allocate a constrained
amount of power across many parallel channels in order to maximize
the sum-rate. Assuming that the power-rate function for each channel
is proportional to $\log(1+SNR)$ as per the Shannon's capacity
theorem for AWGN channels, it is well known that the optimal power
allocation can be determined by a water-filling
strategy~\cite{Cover:1991}. The classic water-filling solution is a
deterministic algorithm, and requires perfect knowledge of all
channel gain to noise ratios.

In practice, however, channel gain-to-noise ratios are stochastic
quantities. To handle this randomness, we consider an alternative
approach, based on online learning, specifically stochastic
multi-armed bandits. We formulate the problem of stochastic
water-filling as follows: time is discretized into slots; each
channel's gain-to-noise ratio is modeled as an i.i.d. random
variable with an unknown distribution. In our general formulation,
the power-to-rate function for each channel is allowed to be any
sub-additive function~\footnote{A function $f$ is subadditive if
$f(x + y) \le f(x)+ f(y)$; for any concave function $g$, if $g(0)
\geq 0$ (such as $\log(1+x)$), $g$ is subadditive.}. We seek a power
allocation that maximizes the expected sum-rate (i.e., an
optimization of the form $\mathbb{E}[\sum\limits_i \log(1+SNR_i)]$).
Even if the channel gain-to-noise ratios are random variables with
known distributions, this turns out to be a hard combinatorial
stochastic optimization problem. Our focus in this paper is thus on
a more challenging case.

In the classical multi-armed bandit, there is a player playing $K$
arms that yield stochastic rewards with unknown means at each time
in i.i.d. fashion over time. The player seeks a policy to maximize
its total expected reward over time. The performance metric of
interest in such problems is regret, defined as the cumulative
difference in expected reward between a model-aware genie and that
obtained by the given learning policy. And it is of interest to show
that the regret grows sub-linearly with time so that the
time-averaged regret asymptotically goes to zero, implying that the
time-averaged reward of the model-aware genie is obtained
asymptotically by the learning policy.

We show that it is possible to map the problem of stochastic
water-filling to an MAB formulation by treating each possible power
allocation as an arm (we consider discrete power levels in this
paper; if there are $P$ possible power levels for each of $N$
channels, there would be $P^N$ total arms.) We present a novel
combinatorial policy for this problem that we call CWF1, that yields
regret growing polynomially in $N$ and logarithmically over time.
Despite the exponential growing set of arms, the CWF1 observes and
maintains information for $P\cdot N$ variables, one corresponding to
each power-level and channel, and exploits linear dependencies
between the arms based on these variables.

Typically, the way the randomness in the channel gain to noise
ratios is dealt with is that the mean channel gain to noise ratios
are estimated first based on averaging a finite set of training
observations and then the estimated gains are used in a
deterministic water-filling procedure. Essentially this approach
tries to identify the power allocation that maximizes a
pseudo-sum-rate, which is determined based on the power-rate
equation applied to the mean channel gain-to-noise ratios (i.e., an
optimization of the form $\sum\limits_i \log(1 +
\mathbb{E}[SNR_i]$). We also present a different stochastic
water-filling algorithm that we call CWF2, which learns to do this
in an online fashion. This algorithm observes and maintains
information for $N$ variables, one corresponding to each channel,
and exploits non-linear dependencies between the arms based on these
variables. To our knowledge, CWF2 is the first MAB algorithm to
exploit non-linear dependencies between the arms. We show that the
number of times CWF2 plays a non-optimal combination of powers is
uniformly bounded by a function that is logarithmic in time. Under
some restrictive conditions, CWF2 may also solve the first problem
more efficiently.

\section{Related Work}\label{sec:related}

The classic water-filling strategy is described in
~\cite{Cover:1991}. There are a few other stochastic variations of
water-filling that have been covered in the literature that are
different in spirit from our formulation. When a fading distribution
over the gains is known \emph{a priori}, the power constraint is
expressed over time, and \emph{the instantaneous gains are also
known}, then a deterministic joint frequency-time water-filling
strategy can be used~\cite{GoldsmithVaraiya, GoldsmithBook}.
In~\cite{Wang:2010}, a stochastic gradient approach based on
Lagrange duality is proposed to solve this problem when the fading
distribution is unknown but still instantaneous gains are available.
By contrast, in our work we do not assume that the instantaneous
gains are known, and focus on keeping the same power constraint at
each time while considering unknown gain distributions.

%http://www.ece.utexas.edu/~jandrews/publications/SheHea_Globecom04.pdf

Another work~\cite{Zaidi:2005} considers water-filling over
stochastic non-stationary fading channels, and proposes an adaptive
learning algorithm that tracks the time-varying optimal power
allocation by incorporating a forgetting factor. However, the focus
of their algorithm is on minimizing the maximum mean squared error
assuming imperfect channel estimates, and they prove only that their
algorithm would converge in a stationary setting. Although their
algorithm can be viewed as a learning mechanism, they do not treat
stochastic water-filling from the perspective of multi-armed
bandits, which is a novel contribution of our work. In our work, we
focus on stationary setting with perfect channel estimates, but
prove stronger results, showing that our learning algorithm not only
converges to the optimal allocation, it does so with sub-linear
regret.

There has been a long line of work on stochastic multi-armed bandits
involving playing arms yielding stochastically time varying rewards
with unknown distributions. Several authors~\cite{Gai:LLR,
Anantharam, Agrawal:1995, Auer:2002} present learning policies that
yield regret growing logarithmically over time (asymptotically, in
the case of ~\cite{Gai:LLR, Anantharam, Agrawal:1995} and uniformly
over time in the case of~\cite{Auer:2002}). Our algorithms build on
the UCB1 algorithm proposed in~\cite{Auer:2002} but make significant
modifications to handle the combinatorial nature of the arms in this
problem. CWF1 has some commonalities with the LLR algorithm we
recently developed for a completely different problem, that of
stochastic combinatorial bipartite matching for channel
allocation~\cite{Gai:2010}, but is modified to account for the
non-linear power-rate function in this paper. Other recent work on
stochastic MAB has considered decentralized
settings~\cite{Anandkumar:Infocom:2010, Anandkumar:JSAC,
Liu:zhao:2010, Gai:decentralized:globecom}, and non-i.i.d. reward
processes ~\cite{Tekin:2010, Tekin:restless:infocom, Qing:ita,
Dai:icassp, Gai:rested:globecom}. With respect to this literature,
the problem setting for stochastic water-filling is novel in that it
involves a non-linear function of the action and unknown variables.
In particular, as far as we are aware, our CWF2 policy is the first
to exploit the non-linear dependencies between arms to provably
improve the regret performance.

\section{Problem Formulation}\label{sec:formulation}

We define the stochastic version of the classic communication theory
problem of power allocation for maximizing rate over parallel
channels (water-filling) as follows.

We consider a system with $N$ channels, where the channel
gain-to-noise ratios are unknown random processes $X_i(n), 1 \leq i
\leq N $. Time is slotted and indexed by $n$. We assume that
$X_i(n)$ evolves as an i.i.d. random process over time (i.e., we
consider block fading), with the only restriction that its
distribution has a finite support. Without loss of generality, we
normalize $X_i(n) \in [0,1]$. We do not require that $X_i(n)$ be
independent across $i$. This random process is assumed to have a
mean $\theta_{i} = \mathds{E}[X_i]$ that is unknown to the users. We
denote the set of all these means by $\Theta = \{\theta_{i}\}$.

At each decision period $n$ (also referred to interchangeably as a
time slot), an $N$-dimensional action vector $\xa(n)$, representing
a power allocation on these $N$ channels, is selected under a policy
$\pi(n)$. We assume that the power levels are discrete, and we can
put any constraint on the selections of power allocations such that
they are from a finite set $\xF$ (i.e., the maximum total power
constraint, or an upper bound on the maximum allowed power per
subcarrier). We assume $a_i(n) \geq 0$ for all $1 \leq i \leq N$.
When a particular power allocation $\xa(n)$ is selected, the channel
gain-to-noise ratios corresponding to nonzero components of $\xa(n)$
are revealed, i.e., the value of $X_i(n)$ is observed for all $i$
such that $a_i(n) \neq 0$. We denote by $\xA_{\xa(n)} = \{i: a_i(n)
\neq 0, 1 \leq i \leq N \}$ the index set of all $a_i(n) \neq 0$ for
an allocation $\xa$.

We adopt a general formulation for water-filling, where the sum rate
\footnote{We refer to rate and reward interchangeably in this
paper.} obtained at time $n$ by allocating a set of powers $\xa(n)$
is defined as:
\begin{equation}
  R_{\xa(n)}(n) = \sum\limits_{i \in \xA_{\xa(n)}} f_i(a_i(n), X_i(n)).
\end{equation}
where for all $i$, $f_i(a_i(n), X_i(n))$ is a nonlinear continuous
increasing sub-additive function in $X_i(n)$, and $f_i(a_i(n), 0) =
0$ for any $a_i(n)$. We assume $f_i$ is defined on
$\mathds{R}^{+}\times \mathds{R}^{+}$.

Our formulation is general enough to include as a special case of
the rate function obtained from Shannon's capacity theorem for AWGN,
which is widely used in communication networks:
\begin{equation}
   R_{\xa(n)}(n) = \sum\limits_{i = 1}^N \log( 1 + a_i(n)
  X_i(n)) \nonumber
\end{equation}
In the typical formulation there is a total power constraint and
individual power constraints, the corresponding constraint is
\begin{equation}
  \xF = \{\xa:  \sum\limits_{i = 1}^N a_i \leq P_{\text{total}} \wedge 0 \leq a_i \leq
  P_i, \forall i\}. \nonumber
\end{equation}
where $P_{\text{total}}$ is the total power constraint and $P_i$ is
the maximum allowed power per channel.

%We consider two objectives in this research.
%
%\subsection{Maximizing the Expected
%Sum-Rate}

Our goal is to maximize the expected sum-rate when the distributions
of all $X_i$ are unknown, as shown in (\ref{equ:o1}). We refer to
this objective as $\mathbf{O_1}$.
\begin{equation} \label{equ:o1}
 \max\limits_{\xa \in\xF}  \mathds{E}[\sum \limits_{i \in \xA_{\xa}} f_i(a_i,
 X_i))]
\end{equation}

Note that even when $X_i$ have known distributions, this is a hard
combinatorial non-linear stochastic optimization problem. In our
setting, with unknown distributions, we can formulate this as a
multi-armed bandit problem, where each power allocation $\xa(n) \in
\xF$ is an arm and the reward function is in a combinatorial
non-linear form. The optimal arms are the ones with the largest
expected reward, denoted as $\mathcal{O}^* = \{\xa^* \}$. For the
rest of the paper, we use $*$ as the index indicating that a
parameter is for an optimal arm. If more than one optimal arm
exists, $*$ refers to any one of them.

We note that for the combinatorial multi-armed bandit problem with
linear rewards where the reward function is defined by
$R_{\xa(n)}(n) = \sum\limits_{i \in \xA_{\xa(n)}} a_i(n) X_i(n)$,
$\xa^*$ is a solution to a deterministic optimization problem
because $\max\limits_{\xa \in\xF} \mathds{E}[\sum\limits_{i \in
\xA_{\xa}} a_i X_i] = \max\limits_{\xa \in\xF} \sum\limits_{i \in
\xA_{\xa}} a_i \mathds{E}[X_i]$. Different from the combinatorial
multi-armed bandit problem with linear rewards, $\xa^*$ here is a
solution to a stochastic optimization problem, i.e.,
\begin{equation}
 \xa^* \in \mathcal{O}^* = \{\tilde{\xa}: \tilde{\xa} = \arg\max\limits_{\xa \in\xF} \mathds{E}[\sum \limits_{i \in \xA_{\xa}} f_i(a_i,
 X_i))] \}.
\end{equation}

We evaluate policies for $\mathbf{O_1}$ with respect to
\emph{regret}, which is defined as the difference between the
expected reward that could be obtained by a genie that can pick an
optimal arm at each time, and that obtained by the given policy.
Note that minimizing the regret is equivalent to maximizing the
expected rewards. Regret can be expressed as:
\begin{equation}
 \mathfrak{R}^\pi (n) = n R^*  - \mathds{E}[ \sum \limits_{t = 1}^n R_{\pi(t)}(t) ],
\end{equation}
where $R^* = \max\limits_{\xa \in\xF} \mathds{E}[\sum\limits_{i \in
\xA_{\xa}} f_i(a_i, X_i))]$, the expected reward of an optimal arm.

Intuitively, we would like the regret $\mathfrak{R}^\pi (n)$ to be
as small as possible. If it is sub-linear with respect to time $n$,
the time-averaged regret will tend to zero and the maximum possible
time-averaged reward can be achieved. Note that the number of arms
$|\xF|$ can be exponential in the number of unknown random variables
$N$.

%\subsection{??}

We also note that for the stochastic version of the water-filling
problems, a typical way in practice to deal with the unknown
randomness is to estimate the mean channel gain to noise ratios
first and then find the optimized allocation based on the mean
values. This approach tries to identify the power allocation that
maximizes the power-rate equation applied to the mean channel
gain-to-noise ratios. We refer to maximizing this as the
sum-pseudo-rate over averaged channels. We denote this objective by
$\mathbf{O_2}$, as shown in (\ref{equ:o2}).
\begin{equation} \label{equ:o2}
 \max\limits_{\xa \in\xF}  \sum \limits_{i \in \xA_{\xa}} f_i(a_i,
 \mathds{E}[X_i])
\end{equation}

We would also like to develop an online learning policy for
$\mathbf{O_2}$. Note that the optimal arm $\xa^*$ of $\mathbf{O_2}$
is a solution to a deterministic optimization problem. So, we
evaluate the policies for $\mathbf{O_2}$ with respect to the
expected total number of times that a non-optimal power allocation
is selected. We denote by $T_{\xa}(n)$ the number of times that a
power allocation is picked up to time $n$.  We denote $r_{\xa} =
\sum\limits_{i \in \xA_{\xa}} f_i(a_i, \mathds{E}[X_i])$. Let
$T_{non}^{\pi}(n)$ denote the total number of times that a policy
$\pi$ select a power allocation $r^{\xa} < r^{\xa^*}$. Denote by
$\mathds{1}_t^{\pi} (\xa)$ the indicator function which is equal to
$1$ if $\xa$ is selected under policy $\pi$ at time $t$, and 0 else.
Then
\begin{align}
  \mathbb{E}[T_{non}^{\pi} (n)] & = n - \mathbb{E}[\sum\limits_{t = 1}^n
  \mathds{1}_t^{\pi} (\xa^*) = 1] \\
  & = \sum\limits_{r_{\xa} < r_{\xa^*}}
  \mathbb{E}[T_{\xa}(n)].\nonumber
\end{align}

%Generally speaking, $\mathds{E}[\sum \limits_{i \in \xA_{\xa}}
%f_i(a_i, X_i))] \neq \sum \limits_{i \in \xA_{\xa}} f_i(a_i,
%\mathds{E}[X_i]))$ and this results more challenges for this
%problem.

\section{Online Learning for Maximizing the Sum-Rate }\label{sec:general}

We first present in this section an online learning policy for
stochastic water-filling under object $\mathbf{O_1}$.

\subsection{Policy Design}\label{subsec:general:policy}

A straightforward, naive way to solve this problem is to use the
UCB1 policy proposed \cite{Auer:2002}. For UCB1, each power
allocation is treated as an arm, and the arm that maximizes
$\hat{Y}_k + \sqrt{ \frac{2 \ln n }{ m_k } }  $ will be selected at
each time slot, where $\hat{Y}_k$ is the mean observed reward on arm
$k$, and $m_k$ is the number of times that arm $k$ has been played.
This approach essentially ignores the underlying dependencies across
the different arms, and requires storage that is linear in the
number of arms and yields regret growing linearly with the number of
arms. Since there can be an exponential number of arms, the UCB1
algorithm performs poorly on this problem.

We note that for combinatorial optimization problems with linear
reward functions, an online learning algorithm LLR has been proposed
in \cite{Gai:LLR} as an efficient solution. LLR stores the mean of
observed values for every underlying unknown random variable, as
well as the number of times each has been observed. So the storage
of LLR is linear in the number of unknown random variables, and the
analysis in \cite{Gai:LLR} shows LLR achieves a regret that grows
logarithmically in time, and polynomially in the number of unknown
parameters.

However, the challenge with stochastic water-filling with objective
$\mathbf{O_1}$, where the expectation is outside the non-linear
reward function, directly storing the mean observations of $X_i$
will not work.

To deal with this challenge, we propose to store the information for
each $a_i, X_i$ combination, i.e., $\forall 1\leq i \leq N$,
$\forall a_i$, we define a new set of random variables $Y_{i,a_i} =
f_i(a_i, X_i)$. So now the number of random variables $Y_{i,a_i}$ is
$\sum\limits_{i = 1}^N |\mathcal{B}_i|$, where $\mathcal{B}_i =
\{a_i: a_i \neq 0\}$. Note that $\sum\limits_{i = 1}^N
|\mathcal{B}_i| \leq PN$.

Then the reward function can be expressed as
\begin{equation}\label{equ:30}
  R_{\xa} = \sum\limits_{i \in \xA_{\xa}} Y_{i,a_i},
\end{equation}
Note that (\ref{equ:30}) is in a combinatorial linear form.

For this redefined MAB problem with $\sum\limits_{i = 1}^N
|\mathcal{B}_i|$ unknown random variables and linear reward function
(\ref{equ:30}), we propose the following online learning policy CWF1
for stochastic water-filling as shown in Algorithm \ref{alg:CWF1}.

%%%%%%%%%%%%%%%%%%%%%%
%%%%%%%%%%%%%%%%%%%%%%%
\begin{algorithm} [ht]
\caption{Online Learning for Stochastic Water-Filling: CWF1 }
\label{alg:CWF1}

\begin{algorithmic}[1]
%----------------------------------------------------------------
\State \commnt{ Initialization}

\State If $\max\limits_\xa |\xA_\xa|$ is known, let $L =
\max\limits_\xa |\xA_\xa|$; else, $L = N$;

\For {$n = 1$ to $N$}
    \State Play any arm $\xa$ such that $n \in \xA_\xa$;
    \State $\forall i \in \xA_{\xa}$, $\forall a_i \in \mathcal{B}_i$, $\overline{Y}_{i,a_i} := \frac{\overline{Y}_{i,a_i} m_{i} + f_i(a_i, X_{i})}{ m_{i}
    +1}$;
    \State $\forall i \in \xA_{\xa}$, $m_{i} := m_{i}+1$;
\EndFor

\State \commnt{Main loop}

\While {1}
    \State $n := n + 1$;
    \State Play an arm $\xa $ which solves the maximization problem
    \begin{equation}
    \label{equ:choosemax}
     \sum\limits_{i \in \xA_{\xa}} (\overline{Y}_{i,a_i} + \sqrt{ \frac{ (L+1) \ln n }{
    m_{i}}} );
    \end{equation}
    \State $\forall i \in \xA_{\xa}$, $\forall a_i \in \mathcal{B}_i$, $\overline{Y}_{i,a_i} := \frac{\overline{Y}_{i,a_i} m_{i} + f_i(a_i, X_{i})}{ m_{i}
    +1}$; \label{line:12}
    \State $\forall i \in \xA_{\xa}$, $m_{i} := m_{i}+1$;
\EndWhile
\end{algorithmic}
\end{algorithm}
%%%%%%%%%%%%%%%%%%%%%%%
%%%%%%%%%%%%%%%%%%%%%%%

To have a tighter bound of regret, different from the LLR algorithm,
instead of storing the number of times that each unknown random
variables $Y_{i,a_i}$ has been observed, we use a $1$ by $N$ vector,
denoted as $(m_{i})_{1 \times N}$, to store the number of times that
$X_i$ has been observed up to the current time slot.

We use a $1$ by $\sum\limits_{i = 1}^N |\mathcal{B}_i|$ vector,
denoted as $(\overline{Y}_{i,a_i})_{1 \times \sum\limits_{i = 1}^N
|\mathcal{B}_i|}$ to store the information based on the observed
values. $(\overline{Y}_{i,a_i})_{1 \times \sum\limits_{i = 1}^N
|\mathcal{B}_i|}$ is updated in as shown in line \ref{line:12}.
% store the average of all the observed values
%(calculated based on the observed value of $X_i$) of $Y_i(a_i)$ up
%to the current time slot.
Each time an arm $\xa(n)$ is played, $\forall i \in \xA_{\xa(n)}$,
the observed value of $X_i$ is obtained. For every observed value of
$X_i$, $|\mathcal{B}_i|$ values are updated: $\forall a_i \in
\mathcal{B}_i$, the average value $\overline{Y}_{i,a_i}$ of all the
values of $Y_{i,a_i}$ up to the current time slot is updated. CWF1
policy requires storage linear in $\sum\limits_{i = 1}^N
|\mathcal{B}_i|$.

%Different from the LLR algorithm, we use one $1$ by $\sum\limits_{i
%= 1}^N |\mathcal{B}_i|$ vector, denoted as $(\overline{Y_i(a_i)})_{1
%\times \sum\limits_{i = 1}^N |\mathcal{B}_i|}$ to store the average
%of all the observed values of $Y_i(a_i)$ up to the current time
%slot. We use another  $1$ by $N$ vector, denoted as $(m_{i})_{1
%\times N}$, to store the number of times that $X_i$ has been
%observed up to the current time slot.

\subsection{Analysis of regret}\label{subsec:general:regret}

%The upper bound of regret for LLR policy has been shown in
%\cite{Gai:LLR}. We rewrite the theorem as below.
%
%\begin{theorem}\label{c1:upperbound}[Theorem 2 in \cite{Gai:LLR}] The expected
%regret under the LLR policy is at most
%\begin{equation} \label{equ:upperbound}
%\left[\frac{ 4 a_{\max}^2 L^2 (L+1) N \ln n }{ \left( \Delta_{\min}
%\right)^2 } + N + \frac{\pi^2}{3} L N \right] \Delta_{\max}.
%\end{equation}
%\end{theorem}

%For CWF1 policy, although there are $\sum\limits_{i = 1}^N
%|\mathcal{B}_i|$ random variables, the upper bound of regret remain
%the same, as in Theorem \ref{c1:mllr}.

\theorem\label{c1:mllr} The expected regret under the CWF1 policy is
at most
\begin{equation} \label{equ:upperbound}
%\mathfrak{R}^{CWF1} (n) \leq
\left[\frac{ 4 a_{\max}^2 L^2 (L+1) N
\ln n }{ \left( \Delta_{\min} \right)^2 } + N + \frac{\pi^2}{3} L N
\right] \Delta_{\max}.
\end{equation}
where $a_{\max} = \max\limits_{\xa \in \xF} \max\limits_{i} a_i$,
$\Delta_{\min} = \min\limits_{\xa \neq \xa^*} R^* -
\mathbb{E}[R_{\xa}]$, $\Delta_{\max} = \max\limits_{\xa \neq \xa^*}
R^* - \mathbb{E}[R_{\xa}]$. Note that $L \leq N$.

%\begin{IEEEproof}
%  See \cite{Gai:waterfilling}.
%\end{IEEEproof}

The proof of Theorem \ref{c1:mllr} is omitted.

\begin{remark} For CWF1 policy, although there are $\sum\limits_{i = 1}^N
|\mathcal{B}_i|$ random variables, the upper bound of regret remains
$O(N^4 \log n)$, which is the same as LLR, as shown by Theorem 2 in
\cite{Gai:LLR}. Directly applying LLR algorithm to solve the
redefined MAB problem in (\ref{equ:30}) will result in a regret that
grows as $O(P^4 N^4 \log n)$.
\end{remark}

\begin{remark} Algorithm \ref{alg:CWF1} will even work for rate functions
that do not satisfy subadditivity.
\end{remark}

\begin{remark} We can develop similar policies and results when $X_i$ are
Markovian rewards as in \cite{Gai:rested:globecom} and
\cite{Gai:restless:techreport}.
\end{remark}

\section{Online Learning for Sum-Pseudo-Rate}\label{sec:concave}

We now show our novel online learning algorithm CWF2 for stochastic
water-filling with object $\mathbf{O_2}$. Unlike CWF1, CWF2 exploits
non-linear dependencies between the choices of power allocations and
requires lower storage. Under condition where the power allocation
that maximize $\mathbf{O_2}$ also maximize $\mathbf{O_1}$, we will
see through simulations that CWF2 has better regret performance.

%\begin{definition}A function $f: A \rightarrow B$ is \emph{subadditive} if
%\begin{equation}
%    \forall x, y \in A, f(x+y) \leq f(x)+f(y).
%\end{equation}
%\end{definition}

%We define (***OPT) if $\exists \xa^* \in \mathcal{O}^*$, such that
%$\forall \xa \notin \mathcal{O}^*$,
%\begin{equation} \label{equ:cond1}
%  \sum\limits_{i \in \xA_{\xa^*}} f_i(a_i, \theta_i)) > \sum\limits_{j \in \xA_{\xa}} f_j(a_j, \theta_j).
%\end{equation}
%For the above problems, we define $\delta_{\xa} = \sum\limits_{i \in
%\xA_{\xa^*}} f_i(a_i,
% \theta_i)) - \sum\limits_{j \in \xA_{\xa}} f_j(a_j, \theta_j)$, $\delta_{\min} = \min\limits_{\xa \notin \mathcal{O}^*}
% \delta_{\xa}$.

\subsection{Policy Design}\label{subsec:concave:policy}

Our proposed policy CWF2 for stochastic water filling with objective
$\mathbf{O_2}$ is shown in Algorithm \ref{alg:concave}.

%%%%%%%%%%%%%%%%%%%%%%
%%%%%%%%%%%%%%%%%%%%%%%
\begin{algorithm} [ht]
\caption{Online Learning for Stochastic Water-Filling: CWF2 }
\label{alg:concave}

\begin{algorithmic}[1]
%----------------------------------------------------------------
\State \commnt{ Initialization}

\State If $\max\limits_\xa |\xA_\xa|$ is known, let $L =
\max\limits_\xa |\xA_\xa|$; else, $L = N$;

\For {$n = 1$ to $N$}
    \State Play any arm $\xa$ such that $n \in \xA_{\xa}$;
    \State $\forall i \in \xA_{\xa}$, $\aX_{i}:= \frac{\aX_{i} m_{i} + X_{i}}{ m_{i}
    +1}$,
     $m_{i} := m_{i}+1$;
\EndFor

\State \commnt{Main loop}

\While {1}
    \State $n := n + 1$;
    \State Play an arm $\xa$ which solves the maximization problem
    \begin{equation}
    \label{equ:choosemax}
    \max\limits_{\xa \in \xF} \sum\limits_{i \in \xA_\xa} \left(f_i(a_i, \aX_{i}) + f_i (a_i, \sqrt{ \frac{ (L+1) \ln n }{
    m_{i}}}) \right);
    \end{equation}
    \State $\forall i \in \xA_{\xa(n)}$, $\aX_{i}:= \frac{\aX_{i} m_{i} + X_{i}}{ m_{i}
    +1}$,
     $m_{i} := m_{i}+1$;
%    \State $\forall i \in \xA_{\xa(n)}$, $\aX_{i}(n) = \frac{\aX_{i}(n-1) m_{i}(n-1) + X_{i}(n)}{ m_{i}(n-1)
%    +1}$, $\quad\quad$
%     $m_{i}(n) = m_{i}(n-1)+1$;
\EndWhile
\end{algorithmic}
\end{algorithm}
%%%%%%%%%%%%%%%%%%%%%%%
%%%%%%%%%%%%%%%%%%%%%%%

We use two $1$ by $N$ vectors to store the information after we play
an arm at each time slot. One is $(\aX_{i})_{1 \times N}$ in which
$\aX_{i}$ is the average (sample mean) of all the observed values of
$X_i$ up to the current time slot (obtained through potentially
different sets of arms over time). The other one is $(m_{i})_{1
\times N}$ in which $m_{i}$ is the number of times that $X_i$ has
been observed up to the current time slot. So CWF2 policy requires
storage linear in $N$.

%We also note that the computation time of CWF2 is polynomial, since
%(\ref{equ:choosemax})

%Note that while we indicate the time index in Algorithm
%\label{alg:concave} for notational clarity, it is not necessary to
%store the matrices from previous time steps while running the
%algorithm. So (***) policy requires storage linear in $N$.

\subsection{Analysis of regret}\label{subsec:concave:regret}
For the analysis of the upper bound for $\mathbb{E}[T_{non}^{\pi}
(n)]$ of CWF2 policy, we use the inequalities as stated in the
Chernoff-Hoeffding bound as follows:

\lemma[Chernoff-Hoeffding bound \cite{Pollard}] \label{lemma:1}
$X_1, \ldots, X_n$ are random variables with range $[0,1]$, and
$E[X_t|X_1, \ldots, X_{t-1}] = \mu$, $\forall 1 \leq t \leq n$.
Denote $S_n = \sum X_i$. Then for all $a \geq 0$
\begin{equation}
 \begin{split}
  \mathds{P}\{S_n \geq n \mu + a\} & \leq e^{-2 a^2 /n}\\
  \mathds{P}\{S_n \leq n \mu - a\} & \leq e^{-2 a^2 /n}
 \end{split}
\end{equation}

\theorem\label{c1:upperbound} Under the CWF2 policy, the expected
total number of times that non-optimal power allocations are
selected is at most
\begin{equation}
\mathbb{E}[T_{non}^{\pi} (n)]  \leq  \frac{N(L+1) \ln n}{ B_{\min}^2
} + N + \frac{\pi^2}{3} LN,
\end{equation}
where $B_{\min}$ is a constant defined by $\delta_{\min}$ and $L$;
$\delta_{\min} = \min\limits_{\xa: r_{\xa} < r^*} (r^* - r_{\xa})$.

\begin{IEEEproof}
%[Proof of Theorem \ref{c1:upperbound}]

We will show the upper bound of the regret in three steps: (1)
introduce a counter $\xT(n)$ (defined as below)  and show its
relationship with the upper bound of the regret; (2) show the upper
bound of $\mathds{E}[\xT(n)]$; (3) show the upper bound of
$\mathbb{E}[T_{non}^{\pi} (n)]  $.

\textit{(1) The counter $\xT(n)$}

After the initialization period, $(\xT(n))_{1 \times N}$ is
introduced as a counter and is updated in the following way: at any
time $n$ when a non-optimal power allocation is selected, find $i$
such that $i = \arg \min\limits_{ j \in \xA_\xa(n) } m_{j}$. If
there is only one such power allocation, $\xT(n)$ is increased by
$1$. If there are multiple such power allocations, we arbitrarily
pick one, say $i'$, and increment $\widetilde{T}_{i'}$ by $1$. Based
on the above definition of $\xT(n)$, each time when a non-optimal
power allocation is selected, exactly one element in $(\xT(n))_{1
\times N}$ is incremented by $1$. So the summation of all counters
in $(\xT(n))_{1 \times N}$ equals to the total number that we have
selected the non-optimal power allocations, as below:
\begin{equation}
 \label{equ:f1}
 \sum\limits_{\xa: R_{\xa} < R^*} \mathbb{E}[T_\xa(n)] = \sum\limits_{i = 1}^N
 \mathbb{E}[\xT(n)].
\end{equation}

We also have the following inequality for $\xT(n)$:
\begin{equation}
 \label{equ:f2}
 \xT(n) \leq m_{i}(n), \forall 1 \leq i \leq N.
\end{equation}

%%%%%%%%%%%%%%%%%%%%%%%%%%%%%%
\textit{(2) show the upper bound of $\mathds{E}[\xT(n)]$}

Let $C_{t, m_i}$ denote $\sqrt{ \frac{ (L+1) \ln t }{ m_i} }$.
Denote by $\xI(n)$ the indicator function which is equal to $1$ if
$\xT(n)$ is added by one at time $n$. Let $l$ be an arbitrary
positive integer. Then, we could get the upper bound of
$\mathds{E}[\xT(n)]$ as shown in (\ref{equ:14}), where $\xa(t)$ is
defined as a non-optimal power allocation picked at time $t$ when
$\xI(t) = 1$. Note that $m_{i} = \min\limits_j \{m_j: \forall j \in
\xA_{\xa(t)}\}$. We denote this power allocation by $\xa(t)$ since
at each time that $\xI(t) = 1$, we could get different selections of
power allocations.

\begin{figure*}[t]
\normalsize \setcounter{equation}{14}
\begin{equation}
\begin{split}
 \mathds{E}[\xT(n)] & = \sum\limits_{t = N+1}^n \mathds{P} \{ \xI(t)=1\}
  \leq l + \sum\limits_{t = N+1}^n \mathds{P} \{ \xI(t)=1 , \xT(t-1) \geq l \} \\
 & \leq l + \sum\limits_{t = N+1}^n \mathds{P}\{ \sum\limits_{j \in \xA_{\xa^*}}
 \left(f_j (a_j^*, \aX_{j, m_j(t-1)}) +
   f_j (a_j^*, C_{t-1, m_j(t-1)}) \right) \\
 & \quad\quad\quad \leq  \sum\limits_{j \in \xA_{\xa(t)}} \left( f_j (a_j(t), \aX_{j,
 m_j(t-1)}) +
  f_j( a_j(t), C_{t-1, m_j(t-1)}) \right), \xT(t-1) \geq l \}.  \label{equ:14}
\end{split}
\end{equation}
\setcounter{equation}{15} \hrulefill \vspace*{4pt}
\end{figure*}

\begin{figure*}[t]
\normalsize \setcounter{equation}{15}
\begin{align}
 \mathds{E}[\xT(n)] & \leq l + \sum\limits_{t = N+1}^n \mathds{P} \{ \min\limits_{0 < m_{h_1}, \ldots, m_{h_{|\xA_{\xa*}|}} < t }   \sum\limits_{j = 1}^{|\xA_{\xa*}|} \left(  f_{h_j}(a_{h_j}^*, \aX_{h_j, m_{h_j}})
  + f_{h_j}(a_{h_j}^*, C_{t-1, m_{h_j}}) \right)   \nonumber\\
 & \quad\quad\quad \leq \max\limits_{l \leq m_{p_1}, \ldots, m_{p_{|\xA_{\xa(t)}|}} < t} \sum\limits_{j = 1}^{|\xA_{\xa(t)}|} \left( f_{p_j}(a_{p_j}(t), \aX_{p_j, m_{p_j}})
   + f_{p_j}( a_{p_j}(t), C_{t-1, m_{p_j}}) \right) \} \nonumber\\
 & \leq l + \sum\limits_{t = 2}^{\infty} \sum\limits_{m_{h_1} = 1}^{t-1} \dots \sum\limits_{m_{h_{|\xA^*|}} = 1}^{t-1}
 \sum\limits_{m_{p_1} = l}^{t-1} \dots \sum\limits_{m_{p_{|\xA_{\xa(t)}|}} = l}^{t-1}
    \mathds{P}\{\sum\limits_{j = 1}^{|\xA_{\xa*}|} \left( f_{h_j}(a_{h_j}^*, \aX_{h_j, m_{h_j}}) + f_{h_j}(a_{h_j}^*, C_{t-1, m_{h_j}})
    \right) \nonumber \\
 & \quad\quad\quad \leq \sum\limits_{j = 1}^{|\xA_{\xa(t)}|} \left( f_{p_j}(a_{p_j}(t), \aX_{p_j, m_{p_j}}) + f_{p_j}( a_{p_j}(t), C_{t-1, m_{p_j}}) \right)
 \} \label{equ:15} \\
 & \leq l + \sum\limits_{t = 2}^{\infty} \sum\limits_{m_{h_1} = 1}^{t-1} \dots \sum\limits_{m_{h_{|\xA^*|}} = 1}^{t-1}
 \sum\limits_{m_{p_1} = l}^{t-1} \dots \sum\limits_{m_{p_{|\xA_{\xa(t)}|}} = l}^{t-1}
  \mathds{P}\{ \text{At least one of the following must hold:} \nonumber\\
  & \quad\quad\quad \sum\limits_{j = 1}^{|\xA_{\xa*}|} f_{h_j}(a_{h_j}^*, \aX_{h_j, m_{h_j}}) \leq  r^* - \sum\limits_{j =
1}^{|\xA_{\xa*}|} f_{h_j}(a_{h_j}^*, C_{t-1, m_{h_j}}),\label{equ:ineq1}\\
  & \quad\quad\quad \sum\limits_{j = 1}^{|\xA_{\xa(t)}|} f_{p_j}(a_{p_j}(t), \aX_{p_j, m_{p_j}})
 \geq r_{\xa(t)} + \sum\limits_{j = 1}^{|\xA_{\xa(t)}|} f_{p_j}( a_{p_j}(t), C_{t-1,
 m_{p_j}}), \label{equ:ineq2}\\
  & \quad\quad\quad r^* < r_{\xa(t)}  + 2 \sum\limits_{j = 1}^{|\xA_{\xa(t)}|} f_{p_j}( a_{p_j}(t), C_{t-1,
  m_{p_j}}) \label{equ:ineq3}\}
\end{align}
\setcounter{equation}{19} \hrulefill \vspace*{4pt}
\end{figure*}

Note that $l \leq \xT(t-1)$ implies, $l \leq \xT(t-1) \leq
m_{j}(t-1), \forall j \in \xA_{\xa(t)}$. So we could get an upper
bound of $\mathds{E}[\xT(n)]$ as shown in (\ref{equ:15}),
(\ref{equ:ineq1}), (\ref{equ:ineq2}),
(\ref{equ:ineq3})\footnote{These equations are on the next page due
to the space limitations.}, where $h_j$ ($1 \leq j \leq
|\xA_{\xa*}|$) represents the $j$-th element in $\xA_{\xa*}$; $p_j$
($1 \leq j \leq |\xA_{\xa(t)}|$) represents the $j$-th element in
$\xA_{\xa(t)}$; $r^* = \sum\limits_{j = 1}^{|\xA_{\xa*}|}
f_{h_j}(a_{h_j}^*, \theta_{h_j}) = \sum\limits_{i \in \xA_{\xa^*}}
f_i(a_i,
 \theta_i))$; $r_{\xa(t)} = \sum\limits_{j = 1}^{|\xA_{\xa(t)}|}
f_{p_j}(a_{p_j}(t), \theta_{p_j}) = \sum\limits_{i \in \xA_{\xa}}
f_i(a_i, \theta_i)$. %Note that $r^*$ and $r_{\xa(t)}$ are different
%from $R^*$ and $R_{\xa(t)}$.

Now we show the upper bound of the probabilities for inequalities
(\ref{equ:ineq1}), (\ref{equ:ineq2}) and (\ref{equ:ineq3})
separately. We first find the upper bound of the probability for
(\ref{equ:ineq1}), as shown in (\ref{equ:18}).
%\begin{align}
% & \mathds{P}\{ \sum\limits_{j = 1}^{|\xA_{\xa*}|} \log (1+a_{h_j}^* \aX_{h_j, m_{h_j}}) \nonumber\\
% & \qquad \leq  r^* - \sum\limits_{j =
%1}^{|\xA_{\xa*}|} \log ( 1 + a_{h_j}^* C_{t-1, m_{h_j}}) \} \nonumber\\
% & = \mathds{P}\{ \sum\limits_{j = 1}^{|\xA_{\xa*}|} \left(\log (1+a_{h_j}^* \aX_{h_j, m_{h_j}}) + \log ( 1 + a_{h_j}^* C_{t-1,
%m_{h_j}}) \right) \nonumber\\
%  & \qquad \leq \sum\limits_{j = 1}^{|\xA_{\xa*}|} \log
%(1+a_{h_j}^*
%\aX_{h_j, m_{h_j}})\}  \nonumber\\
%%\end{align}
%%\begin{align}
% & \leq \sum\limits_{j = 1}^{|\xA_{\xa*}|} \mathds{P}\{ \log (1+a_{h_j}^* \aX_{h_j, m_{h_j}}) + \log ( 1 + a_{h_j}^* C_{t-1,
%m_{h_j}}) \nonumber\\
% & \qquad\qquad \leq \log (1+a_{h_j}^* \aX_{h_j, m_{h_j}}) \} \nonumber\\
%& \leq \sum\limits_{j = 1}^{|\xA_{\xa*}|} \mathds{P}\{ \log
%(1+a_{h_j}^* \aX_{h_j, m_{h_j}} + a_{h_j}^* C_{t-1, m_{h_j}})
%\nonumber \\
%& \qquad\qquad \leq \log
%(1+a_{h_j}^* \aX_{h_j, m_{h_j}})\} \label{equ:16}\\
%& = \sum\limits_{j = 1}^{|\xA_{\xa*}|} \mathds{P}\{ a_{h_j}^*
%\aX_{h_j, m_{h_j}} + a_{h_j}^* C_{t-1, m_{h_j}} \leq a_{h_j}^*
%\aX_{h_j,
%m_{h_j}} \} \label{equ:17} \\
%& = \sum\limits_{j = 1}^{|\xA_{\xa*}|} \mathds{P}\{ \aX_{h_j,
%m_{h_j}} + C_{t-1, m_{h_j}} \leq \aX_{h_j, m_{h_j}} \}
%\label{equ:18}
%\end{align}

\begin{figure*}[!t]
\normalsize \setcounter{equation}{19}
\begin{align}
 & \mathds{P}\{ \sum\limits_{j = 1}^{|\xA_{\xa*}|} f_{h_j}(a_{h_j}^*, \aX_{h_j, m_{h_j}}) \leq  r^* - \sum\limits_{j =
1}^{|\xA_{\xa*}|} f_{h_j}(a_{h_j}^*, C_{t-1, m_{h_j}}) \} \nonumber\\
 & = \mathds{P}\{ \sum\limits_{j = 1}^{|\xA_{\xa*}|} \left( f_{h_j}(a_{h_j}^*, \aX_{h_j, m_{h_j}}) + f_{h_j}(a_{h_j}^*, C_{t-1,
m_{h_j}}) \right) \leq \sum\limits_{j = 1}^{|\xA_{\xa*}|}
f_{h_j}(a_{h_j}^*,
\theta_{h_j})\}  \nonumber\\
 & \leq \sum\limits_{j = 1}^{|\xA_{\xa*}|} \mathds{P}\{ f_{h_j}(a_{h_j}^*, \aX_{h_j, m_{h_j}}) + f_{h_j}(a_{h_j}^*, C_{t-1,
m_{h_j}}) \leq f_{h_j}(a_{h_j}^*, \theta_{h_j}) \} \nonumber\\
& \leq \sum\limits_{j = 1}^{|\xA_{\xa*}|} \mathds{P}\{
f_{h_j}(a_{h_j}^*, \aX_{h_j, m_{h_j}} + C_{t-1, m_{h_j}}) \leq
f_{h_j}(a_{h_j}^*, \theta_{h_j})\} \label{equ:16}\\
%& = \sum\limits_{j = 1}^{|\xA_{\xa*}|} \mathds{P}\{ a_{h_j}^*
%\aX_{h_j, m_{h_j}} + a_{h_j}^* C_{t-1, m_{h_j}} \leq a_{h_j}^*
%\theta_{h_j} \} \label{equ:17} \\
& = \sum\limits_{j = 1}^{|\xA_{\xa*}|} \mathds{P}\{ \aX_{h_j,
m_{h_j}} + C_{t-1, m_{h_j}} \leq \theta_{h_j} \} \label{equ:18}
\end{align}
\setcounter{equation}{21} \hrulefill \vspace*{4pt}
\end{figure*}

Equation (\ref{equ:16}) holds because of lemma \ref{lemma:1}. So
$\forall j$,
\begin{equation}
 \begin{split}
& f_{h_j}(a_{h_j}^*, \aX_{h_j, m_{h_j}} + C_{t-1,
m_{h_j}})\\
& \leq f_{h_j}(a_{h_j}^*, \aX_{h_j, m_{h_j}}) + f_{h_j}(a_{h_j}^*,
C_{t-1, m_{h_j}}).
\end{split}
\end{equation}
(\ref{equ:18}) holds because $\forall i$, $f_i(a_i, X_i)$ is a
non-decreasing function in $X_i$ for any $X_i \geq 0$.

In (\ref{equ:18}), $\forall 1 \leq j \leq {|\xA_{\xa*}|}$, applying
the Chernoff-Hoeffding bound stated in Lemma \ref{lemma:1}, we could
find the upper bound of each item as,
\begin{equation}
\begin{split}
 & \mathds{P}\{\aX_{h_j, m_{h_j}} + C_{t-1, m_{h_j}} \leq \theta_{h_j} \} \nonumber \\
 %& = \mathds{P}\{m_{h_j} \aX_{h_j, m_{h_j}} \leq m_{h_j} \aX_{h_j, m_{h_j}} - m_{h_j} C_{t-1, m_{h_j}} \}\\
 & \leq e^{-2\cdot \frac{1}{m_{h_ij}} \cdot (m_{h_j})^2 \cdot \frac{ (L+1) \ln (t-1) }{ m_{h_j} } }
  = (t-1)^{-2(L+1)}.
\end{split}
\end{equation}
Thus,
\begin{equation}
\begin{split}
 & \mathds{P}\{ \sum\limits_{j =
1}^{|\xA_{\xa*}|} f_{h_j}(a_{h_j}^*, \aX_{h_j, m_{h_j}})  \leq r^* -
\sum\limits_{j = 1}^{|\xA_{\xa*}|} f_{h_j}( a_{h_j}^*, C_{t-1,
m_{h_j}}) \} \\
   & \leq |\xA_{\xa*}|
 t^{-2(L+1)}
  \leq \quad L (t-1)^{-2(L+1)} \label{equ:20}.
\end{split}
\end{equation}

Now we can get the upper bound of the probability for inequality
(\ref{equ:ineq2}), as shown in (\ref{equ:22}).

\begin{figure*}[!t]
\normalsize \setcounter{equation}{23}
\begin{align}
 & \mathds{P}\{ \sum\limits_{j = 1}^{|\xA_{\xa(t)}|} f_{p_j}(a_{p_j}(t), \aX_{p_j, m_{p_j}})
 \geq r_{\xa(t)} + \sum\limits_{j = 1}^{|\xA_{\xa(t)}|} f_{p_j}(a_{p_j}(t), C_{t-1, m_{p_j}})
 \} \nonumber \\
 & = \mathds{P}\{ \sum\limits_{j = 1}^{|\xA_{\xa(t)}|} f_{p_j}(a_{p_j}(t), \aX_{p_j, m_{p_j}})
  \geq  \sum\limits_{j =
1}^{|\xA_{\xa(t)}|} \left(f_{p_j}(a_{p_j}(t), \theta_{p_j}) +
f_{p_j}(a_{p_j}(t), C_{t-1, m_{p_j}}) \right)\} \nonumber \\
 & \leq \sum\limits_{j = 1}^{|\xA_{\xa(t)}|} \mathds{P}\{ f_{p_j}(a_{p_j}(t), \aX_{p_j, m_{p_j}}) \geq f_{p_j}(a_{p_j}(t), \theta_{p_j}) +
f_{p_j}(a_{p_j}(t), C_{t-1, m_{p_j}})\} \nonumber\\
 & \leq \sum\limits_{j = 1}^{|\xA_{\xa(t)}|} \mathds{P}\{ f_{p_j}(a_{p_j}(t), \aX_{p_j, m_{p_j}}) \geq f_{p_j} (a_{p_j}(t), \theta_{p_j} + C_{t-1, m_{p_j}})
 \} \nonumber\\
 & = \sum\limits_{j = 1}^{|\xA_{\xa(t)}|} \mathds{P}\{ \aX_{p_j, m_{p_j}} \geq \theta_{p_j} + C_{t-1, m_{p_j}})  \leq L (t-1)^{-2(L+1)} \label{equ:22}.
\end{align}
\setcounter{equation}{24} \hrulefill \vspace*{4pt}
\end{figure*}

Equation (\ref{equ:22}) holds, following a similar reasoning as used
to derive (\ref{equ:20}).

For all $i$ and given any $a_i$, since $f_i(a_i, x)$ is an
increasing, continuous function in $x$, we could find a constant
$B_i(a_i)$ such that
\begin{equation}
 f_i(a_i, B_i(a_i)) = \frac{\delta_{\min}}{2 L}.
\end{equation}
Denote $B_{\min}(\xa) = \min\limits_{i \in \xA_{\xa}} B_i(a_i)$.
Then $\forall i \in \xA_{\xa}$, we have
\begin{equation}
 f_i(a_i, B_{\min}(\xa)) \leq \frac{\delta_{\min}}{2 L}.
\end{equation}

Note that for $l \geq \left\lceil \frac{(L+1) \ln n}{
B_{\min}^2(\xa(t)) } \right\rceil$,
\begin{equation}
 \label{equ:p4}
\begin{split}
 & r^* - r_{\xa(t)}  - 2 \sum\limits_{j = 1}^{|\xA_{\xa(t)}|} f_{p_j}(a_{p_j}(t), C_{t-1,
  m_{p_j}}) \\
 & = r^* - r_{\xa(t)}  - 2 \sum\limits_{j = 1}^{|\xA_{\xa(t)}|} f_{p_j}(
 a_{p_j}(t),
 \sqrt{ \frac{ (L+1) \ln (t-1) }{ m_{p_j}} }) \\
 & \geq r^* - r_{\xa(t)}  - 2 \sum\limits_{j = 1}^{|\xA_{\xa(t)}|}
 f_{p_j}(a_{p_j}(t),
 \sqrt{ \frac{ (L+1) \ln n }{ l} }) \\
  & \geq r^* - r_{\xa(t)}  - 2 \sum\limits_{j = 1}^{|\xA_{\xa(t)}|}
 f_{p_j}(a_{p_j}(t),
 \sqrt{ \frac{ (L+1) \ln n }{ l} }) \\
   & \geq r^* - r_{\xa(t)}  - 2 \sum\limits_{j = 1}^{|\xA_{\xa(t)}|}
 f_{p_j}(a_{p_j}(t), B_{\min}(\xa(t))) \\
    & \geq \delta_{\xa(t)}  - 2 \sum\limits_{j = 1}^{|\xA_{\xa(t)}|}
 \frac{\delta_{\min}}{2 L} \geq \delta_{\xa(t)} - \delta_{\min} \geq 0.
\end{split}
\end{equation}

%Note that for $l \geq \left\lceil \frac{(L+1) \ln n}{
%\left(\frac{e^{\delta_{\xa(t)}/2L} - 1}{a_{\max}} \right)^2 }
%\right\rceil$,
%\begin{equation}
% \label{equ:p4}
%\begin{split}
% & r^* - r_{\xa(t)}  - 2 \sum\limits_{j = 1}^{|\xA_{\xa(t)}|} f_{p_j}(a_{p_j}(t), C_{t-1,
%  m_{p_j}}) \\
% & = r^* - r_{\xa(t)}  - 2 \sum\limits_{j = 1}^{|\xA_{\xa(t)}|} \log( 1 + a_{p_j}(t)
% \sqrt{ \frac{ (L+1) \ln (t-1) }{ m_{p_j}} }) \\
% & \geq r^* - r_{\xa(t)}  - 2 L \log( 1 + a_{\max}
% \sqrt{ \frac{ (L+1) \ln n }{ l} }) \\
% & \geq R^* - R_{\xa(t)} - \delta_{\xa(t)} = 0.
%\end{split}
%\end{equation}

So (\ref{equ:ineq3}) is false when $l \geq \left\lceil \frac{(L+1)
\ln n}{ B_{\min}^2(\xa(t)) } \right\rceil$. We denote $B_{\min} =
\min\limits_{\xa \in \xF} B_{\min}(\xa(t)))$, and let $l \geq
\left\lceil \frac{(L+1) \ln n}{ B_{\min}^2 } \right\rceil$, then
(\ref{equ:ineq3}) is false for all $\xa(t)$.

Therefore, we get the upper bound of $\mathbb{E}[\xT(n)]$ as in
(\ref{equ:27}).
\begin{figure*}[t]
\normalsize \setcounter{equation}{27}
\begin{align}
 & \mathbb{E}[\xT(n)]  \leq \left\lceil \frac{(L+1) \ln n}{ B_{\min}^2 } \right\rceil
  + \sum\limits_{t = 2}^{\infty} \left( \sum\limits_{m_{h_1} = 1}^{t-1} \dots \sum\limits_{m_{h_{|\xA^*|}} = 1}^{t-1}
 \sum\limits_{m_{p_1} = l}^{t-1} \dots \sum\limits_{m_{p_{|\xA_{\xa(t)}|}} = l}^{t-1}  2 L (t-1)^{-2(L+1)} \right) \nonumber\\
 & \leq  \frac{(L+1) \ln n}{ B_{\min}^2 } + 1 + L \sum\limits_{t = 1}^\infty 2 t^{-2}
  \leq  \frac{(L+1) \ln n}{ B_{\min}^2 } + 1 + \frac{\pi^2}{3}
  L. \label{equ:27}
\end{align}
\setcounter{equation}{28} \hrulefill \vspace*{4pt}
\end{figure*}

%\begin{equation}
% \label{equ:p5}
%\begin{split}
% & \mathbb{E}[\xT(n)]  \leq \left\lceil \frac{4(L+1) \ln n}{
%\left(\frac{\delta_{\min}}{L a_{\max}} \right)^2 } \right\rceil \\
% & + \sum\limits_{t = 2}^{\infty} \left( \sum\limits_{m_{h_1} = 1}^{t-1} \dots \sum\limits_{m_{h_{|\xA^*|}} = 1}^{t-1}
% \sum\limits_{m_{p_1} = l}^{t-1} \dots \sum\limits_{m_{p_{|\xA_{\xa(t)}|}} = l}^{t-1}  2 L (t-1)^{-2(L+1)} \right) \\
% & \leq  \frac{ 4 a_{\max}^2 L^2 (L+1)  \ln n }{ \left( \delta_{\min} \right)^2 } + 1 + L \sum\limits_{t = 1}^\infty 2 t^{-2} \\
% & \leq  \frac{ 4 a_{\max}^2 L^2 (L+1)  \ln n }{ \left( \delta_{\min} \right)^2 } + 1 + \frac{\pi^2}{3} L. \\
%\end{split}
%\end{equation}

\textit{(3) Upper bound of $\mathbb{E}[T_{non}^{\pi} (n)]$}

\begin{equation}
 \label{equ:p6}
\begin{split}
 & \mathbb{E}[T_{non}^{\pi} (n)]   = \sum\limits_{\xa: R_{\xa} < R^*} \mathbb{E}[T_{\xa}(n)] \\
 & = \sum\limits_{i = 1}^N  \mathbb{E}[\xT(n)] \\
% & \leq \left[\sum\limits_{i = 1}^N  \frac{ 4 a_{\max}^2 L^2 (L+1)  \ln n }{ \left( \delta_{\min} \right)^2 } + N + \frac{\pi^2}{3} LN \right] \Delta_{\max}
% \\
 & \leq \frac{N(L+1) \ln n}{ B_{\min}^2 } + N + \frac{\pi^2}{3} LN . \\
\end{split}
\end{equation}
\end{IEEEproof}

\begin{remark}\label{remark:4}
CWF2 can be used to solve the stochastic water-filling with
objective $\mathbf{O_1}$ as well if $\exists \xa^* \in
\mathcal{O}^*$, such that $\forall \xa \notin \mathcal{O}^*$,
\begin{equation} \label{equ:cond1}
  \sum\limits_{i \in \xA_{\xa^*}} f_i(a_i, \theta_i)) > \sum\limits_{j \in \xA_{\xa}} f_j(a_j, \theta_j).
\end{equation}
Then the regret of CWF2 is at most
\begin{equation}
\mathfrak{R}^{CWF2}(n) \leq \left[\frac{N(L+1) \ln n}{ B_{\min}^2 }
+ N + \frac{\pi^2}{3} LN \right] \Delta_{\max},
\end{equation}

\end{remark}

%\lemma \label{lemma:2} For any concave function $f$. If $f(0) \geq
%0$, then $f$ is subadditive.
%
%\begin{remark}
%(A more general case $f(\xa, \mathbf{X})$, how to define?).
%\end{remark}

%\subsection{An Example: Stochastic Water-filling}

%\section{Policy Design: $f_i(a_i, x_i) = a_i g_i(x_i)$ ?? }\label{sec:general}

\section{Applications and Numerical Simulation Results}\label{sec:app}

\subsection{Numerical Results for CWF1}\label{subsec:simulation:cwf1}
We now show the numerical results for CWF2 policy. We consider a
OFDM system with $4$ subcarriers. We assume the bandwidth of the
system is $4$ MHz, and the noise density is $-80$ dBw/Hz. We assume
Rayleigh fading with parameter $\sigma = (2, 0.8, 2.8 0.32)$ for $4$
subcarriers. We consider the following objective for our simulation:
\begin{eqnarray}
  \max & & \mathds{E}\left[\sum\limits_{i = 1}^N \log( 1 + a_i(n)
  X_i(n))\right] \label{equ:38}\\
  s.t. & & \sum\limits_{i = 1}^N a_i(n) \leq P_{\text{total}},
  \forall n \label{equ:39}\\
  & &  a_1 (n) \in \{0, 10, 20, 30\}, \forall n \label{equ:40}\\
  & & a_2 (n) \in \{0, 10, 20, 30\}, \forall n \\
  & & a_3 (n) \in \{0, 10, 20, 30, 40\}, \forall n \\
  & &  a_4 (n) \in \{0, 10, 20\}, \forall n \label{equ:43}
\end{eqnarray}
where $P_{\text{total}} = 60$mW ($17.8$ dBm). The unit for above
power constraints from (\ref{equ:40}) to (\ref{equ:43}) is mW. Note
that (\ref{equ:39}) to (\ref{equ:43}) define the constraint set
$\xF$.

For this scenario, there are $140$ different choices of power
allocations, and the optimal power allocation can be calculated as
$(20, 20, 20, 0)$.

\begin{figure}[t]
\center
\includegraphics[width=0.50\textwidth]{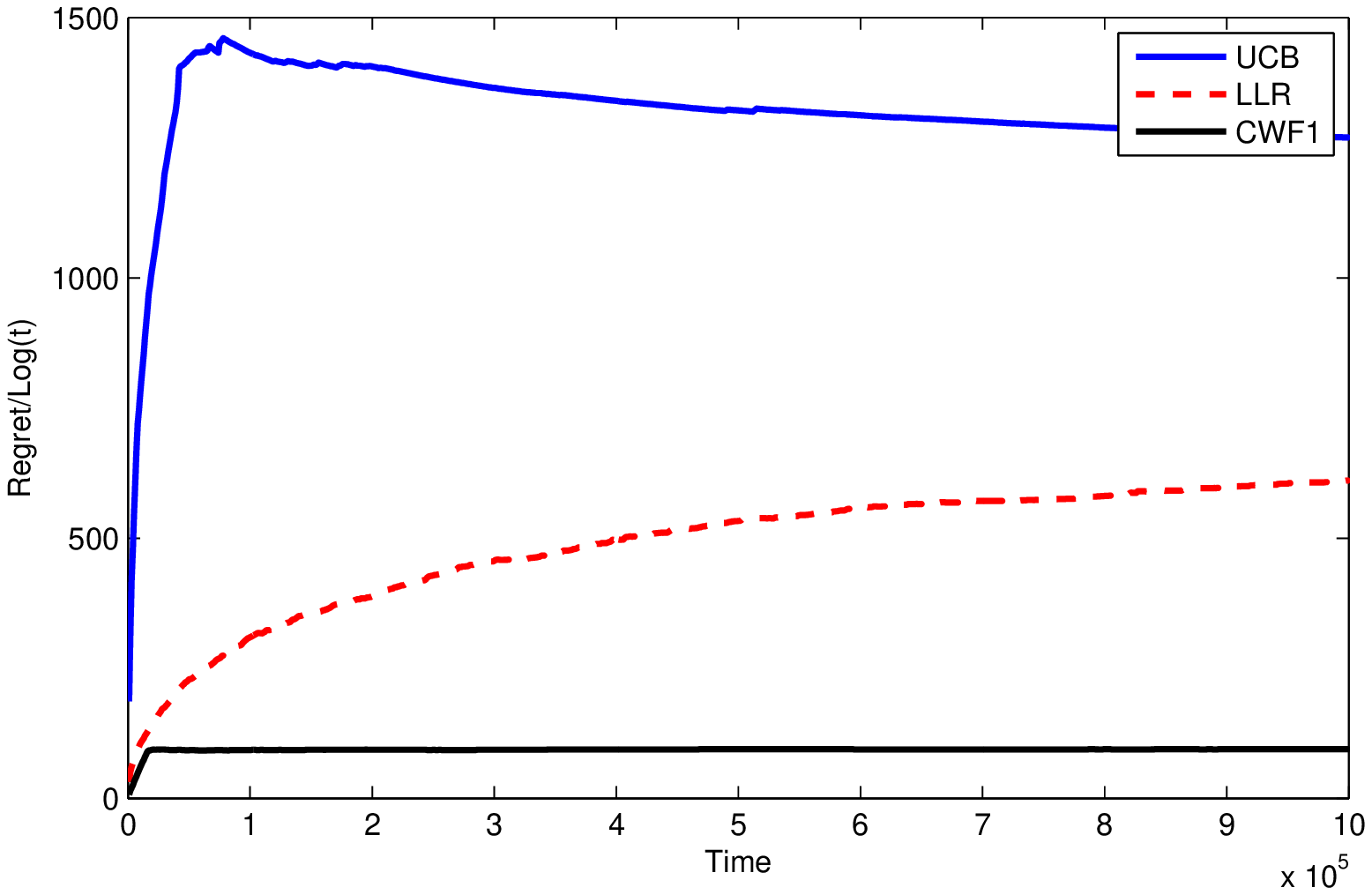}
\caption{Normalized regret $\frac{\mathfrak{R}(n)}{\log n}$ vs. $n$
time slots.} \label{fig:cwf1a}
\end{figure}

We compare the performance of our proposed CWF1 policy with UCB1
policy and LLR policy, as shown in Figure \ref{fig:cwf1a}. As we can
see from \ref{fig:cwf1a}, naively applying UCB1 and LLR policy
results in a worse performance than CWF1, since the UCB1 policy can
not exploit the underlying dependencies across arms, and LLR policy
does not utilize the observations as efficiently as CWF1 does.

\subsection{Numerical Results for CWF2}

We show the simulation results of CWF2 using the same system as in
\ref{subsec:simulation:cwf1}.

We consider the following objective for our simulation:
\begin{eqnarray}
  \max & & \left[\sum\limits_{i = 1}^N \log( 1 + a_i(n)
  \mathds{E}[X_i(n)])\right]\nonumber\\
  s.t. & & \xa \in \xF
\end{eqnarray}
where $\xF$ is same as in \ref{subsec:simulation:cwf1}.

For this scenario, we assume Rayleigh fading with parameter $\sigma
= (1.23, 1.0, 0.55, 0.95)$ for $4$ subcarriers. And the optimal
power allocation can be calculated as $(20, 20, 0, 20)$.

\begin{figure}[t]
\center
\includegraphics[width=0.50\textwidth]{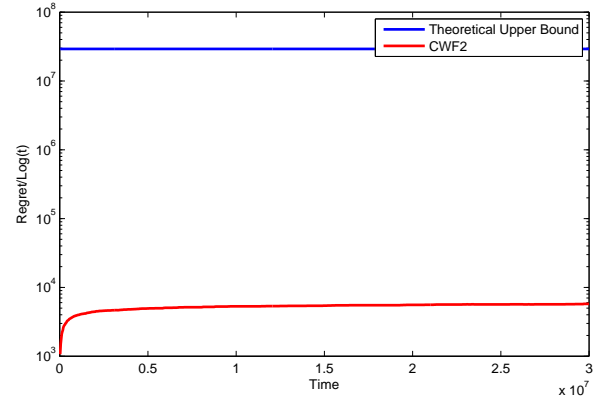}
\caption{Numerical results of $\mathbb{E}[\xT(n)]/\log n$ and
theoretical bound.} \label{fig:cwf2}
\end{figure}

Figure \ref{fig:cwf2} shows the simulation results of the total
number of times that non-optimal power allocations are chosen by
running CWF2 up to 30 million time slots. We also show the
theoretical upper bound in figure \ref{fig:cwf2}. In this case, we
see that the theoretical upper bound is quite loose and the
algorithm does much better in practice.

For this setting, we note that (\ref{equ:cond1}) is satisfied, since
$(20, 20, 0, 20)$ also maximizes (\ref{equ:38}). So as stated in
Remark \ref{remark:4}, CWF2 can also be used to solve stochastic
water filling with $\mathbf{O_1}$, with regret that grows
logarithmically in time and polynomially in the number of channels.

\begin{figure}[t]
\center
\includegraphics[width=0.50\textwidth]{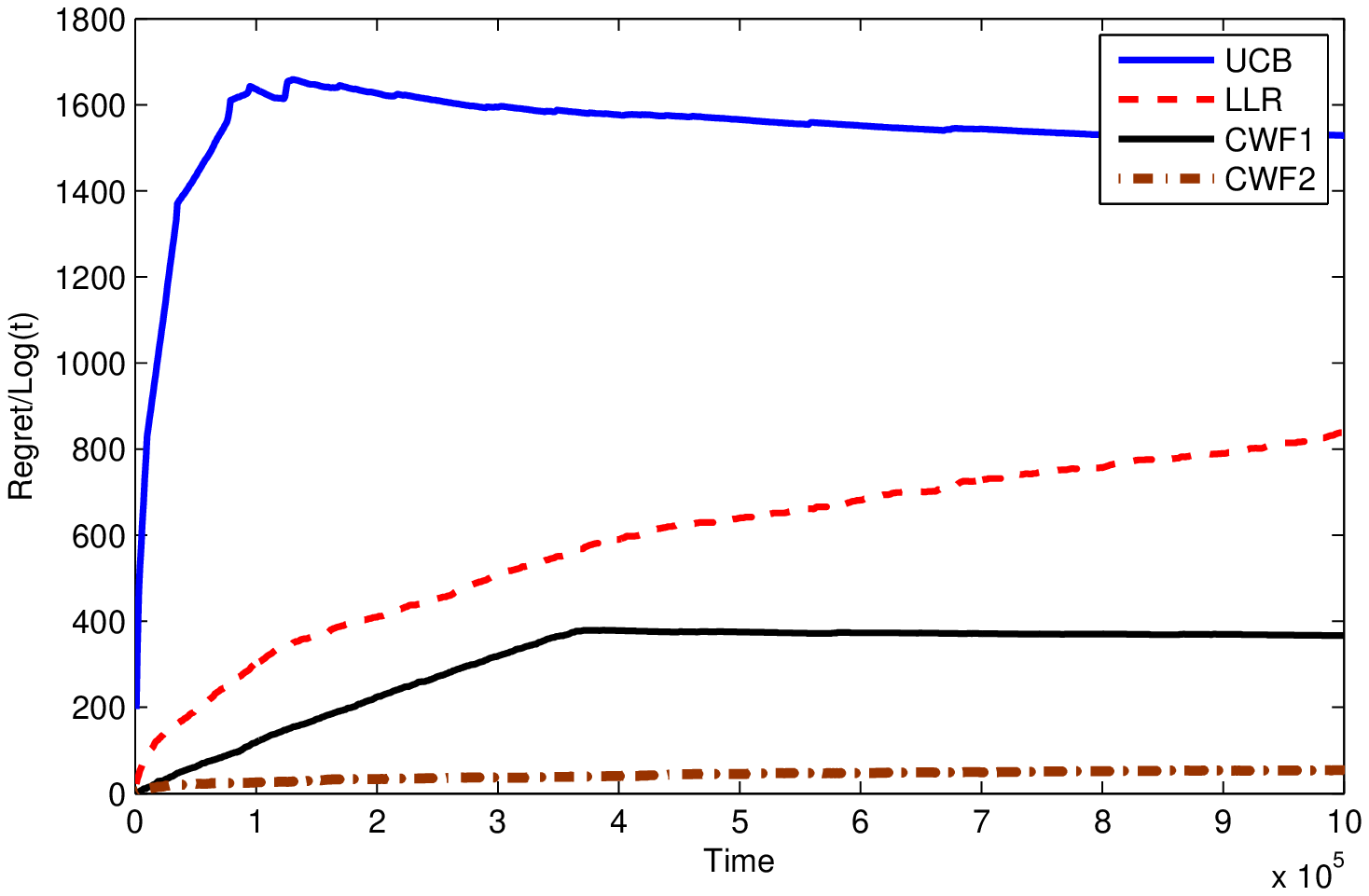}
\caption{Normalized regret $\frac{\mathfrak{R}(n)}{\log n}$ vs. $n$
time slots.} \label{fig:cwf1}
\end{figure}

We show a comparison of the UCB1 policy, LLR policy, CWF1 policy and
CWF2 policy under this setting in Figure \ref{fig:cwf1}. We can see
that CWF2 performs the best by far since it incorporate a way to
exploit non-linear dependencies across arms, and learn more
efficiently.

%As we can see from
%the simulation results, $\mathbb{E}[\xT(n)]/\log n$ tends to flat
%after

%, $P_1 = 30$mW ($14.7$
%dBm), $P_2 = 30$mW, $P_3 = 40$mW ($16$ dbm), $P_4 = 20$mW ($13$
%dbm).

%Table summarizes the parameters for numerical simulation results.
%
%\begin{table}[h]
%\begin{center}
%    \begin{tabular}{ c|c}
%    \hline
%    Parameter Name & Value\\ \hline
%    \hline
%    System bandwidth & $4$ MHz\\
%    Noise PSD & $-60$ dBw/Hz\\
%    Number of subcarriers & 4 \\\hline
%    \end{tabular}
%\vspace{0.1in} \caption{Transition probabilities $p_{01}$, $p_{10}$
%for each user-channel pair} \label{table:2b}
%\end{center}
%\end{table}

\section{Conclusion}\label{sec:conclusion}

We have considered the problem of optimal power allocation over
parallel channels with stochastically time-varying gain-to-noise
ratios for maximizing information rate (stochastic water-filling) in
this work. We approached this problem from the novel perspective of
online learning. The crux of our approach is to map each possible
power allocation into arms in a stochastic multi-armed bandit
problem. The significant new challenge imposed here is that the
reward obtained is a non-linear function of the arm choice and the
underlying unknown random variables. To our knowledge there is no
prior work on stochastic MAB that explicitly treats such a problem.

We first considered the problem of maximizing the expected sum rate.
For this problem we developed the CWF1 algorithm. Despite the fact
that the number of arms grows exponentially in the number of
possible channels, we show that the CWF1 algorithm requires only
polynomial storage and also yields a regret that is polynomial in
the number of power levels per channel and the number of channels,
and logarithmic in time.

We then considered the problem of maximizing the sum-pseudo-rate,
where the pseudo rate for a stochastic channel is defined by
applying the power-rate equation to its mean SNR ($\log(1+E[SNR]$).
The justification for considering this problem is its connection to
practice (where allocations over stochastic channels are made based
on estimated mean channel conditions). Albeit sub-optimal with
respect to maximizing the expected sum-rate, the use of the
sum-pseudo-rate as the objective function is a more tractable
approach. For this problem, we developed a new MAB algorithm that we
call CWF2. This is the first algorithm in the literature on
stochastic MAB that exploits non-linear dependencies between the arm
rewards. We have proved that the number of times this policy uses a
non-optimal power allocation is also bounded by a function that is
polynomial in the number of channels and power-levels, and
logarithmic in time.

Our simulations results show that the algorithms we develop are
indeed better than naive application of classic MAB solutions. We
also see that under settings where the power allocation for
maximizing the sum-pseudo-rate matches the optimal power allocation
that maximizes the expected sum-rate, CWF2 has significantly better
regret-performance than CWF1.

Because our formulations allow for very general classes of
sub-additive reward functions, we believe that our technique may be
much more broadly applicable to settings other than power allocation
for stochastic channels. We would therefore like to identify and
explore such applications in future work.

%\newpage
%\setcounter{page}{16}

\end{document}